\documentclass[conference]{IEEEtran}
\IEEEoverridecommandlockouts
\makeatletter
\def\ps@IEEEtitlepagestyle{%
  \def\@oddfoot{%
    \hfill
    \makebox[\textwidth][c]{%
      \parbox[t]{0.92\textwidth}{%
        \footnotesize
        © 2026 IEEE. Personal use of this material is permitted. Permission from IEEE must be obtained for all other uses, in any current or future media, including reprinting/republishing this material for advertising or promotional purposes, creating new collective works, for resale or redistribution to servers or lists, or reuse of any copyrighted component of this work in other works.%
      }%
    }%
    \hfill
  }%
  \def\@evenfoot{}%
}
\makeatother
\usepackage{cite}
\usepackage{amsmath,amssymb,amsfonts}
\usepackage{algorithmic}
\usepackage{graphicx}
\usepackage{textcomp}
\usepackage{xcolor}
\usepackage{tabularx}
\usepackage{booktabs}
\usepackage{pifont}
\usepackage{graphicx}
\usepackage{float}
\usepackage{placeins}
\usepackage{fontawesome} % Add this to your preamble for lock symbol

\def\BibTeX{{\rm B\kern-.05em{\sc i\kern-.025em b}\kern-.08em
    T\kern-.1667em\lower.7ex\hbox{E}\kern-.125emX}}
\begin{document}

\title{Faster-HEAL: An Efficient and Privacy-Preserving Collaborative Perception Framework for Heterogeneous Autonomous Vehicles}
%\thanks{© 2026 IEEE.  Personal use of this material is permitted.  Permission from IEEE must be obtained for all other uses, in any current or future media, including reprinting/republishing this material for advertising or promotional purposes, creating new collective works, for resale or redistribution to servers or lists, or reuse of any copyrighted component of this work in other works.}

\author{\IEEEauthorblockN{Armin Maleki}
\IEEEauthorblockA{\textit{Electrical and Computer Engineering}\\
\textit{Michigan State University}\\
East Lansing, Michigan\\
malekiar@msu.edu}
\and
\IEEEauthorblockN{Hayder Radha}
\IEEEauthorblockA{\textit{Electrical and Computer Engineering}\\
\textit{Michigan State University}\\
East Lansing, Michigan\\
radha@msu.edu}
}

\maketitle

\begin{abstract}
Collaborative perception (CP) is a promising paradigm for improving situational awareness in autonomous vehicles by overcoming the limitations of single-agent perception. However, most existing approaches assume homogeneous agents, which restricts their applicability in real-world scenarios where vehicles use diverse sensors and perception models. This heterogeneity introduces a feature domain gap that degrades detection performance. Prior works address this issue by retraining entire models/major components, or using feature interpreters for each new agent type, which is computationally expensive, compromises privacy, and may reduce single-agent accuracy. We propose \textbf{Faster-HEAL}, a lightweight and privacy-preserving CP framework that fine-tunes a low-rank visual prompt to align heterogeneous features with a unified feature space while leveraging pyramid fusion for robust feature aggregation. This approach reduces the trainable parameters by $94\%$, enabling efficient adaptation to new agents without retraining large models. Experiments on the OPV2V-H dataset show that \textsc{Faster-HEAL} improves detection performance by $2\%$ over state-of-the-art methods with significantly lower computational overhead, offering a practical solution for scalable heterogeneous CP.
\end{abstract}


\section{Introduction}
With the growing deployment of autonomous vehicles, ensuring reliable and safe operation is critical. Collaborative perception (CP) has emerged as a key paradigm to enhance situational awareness in connected autonomous vehicles (CAVs) by sharing perception information, giving agents a more complete view of occluded and long-range regions and improving safety and accuracy~\cite{b1}. Despite growing interest, collaborative perception still faces bandwidth limits, communication latency, sensor noise, and pose estimation errors, which can severely degrade system performance. While recent work has made progress in mitigating bandwidth, latency, and pose issues~\cite{b2,b3,b4,b6,b7}, most methods assumed homogeneous sensors and identical perception models across vehicles, which simplifies design and limits applicability
in real-world deployments. In practice, vehicles from different manufacturers use diverse sensor configurations and perception networks, creating a feature domain gap between agents. This heterogeneity can degrade downstream perception performance and safety; hence, bridging this gap is essential for robust, safety-critical collaborative perception in real-world autonomous driving.

Prior work on heterogeneous collaborative perception falls into two main approaches. The first closes the domain gap by retraining parts of the ego or neighbor models to project features into a unified feature space~\cite{b8,b9,b11,b12}. Specifically, HEterogeneous ALliance (HEAL)~\cite{b12} uses two stages: first, the ego and \emph{homogeneous agents} are trained with a standard collaborative perception strategy to build a unified feature space, and then each new heterogeneous agent retrains its encoder and other modules to map its inputs into that space. This approach is effective but impractical, since every new agent type must be retrained, which is costly for real-time deployment and can harm both privacy and single-agent performance. Because some vehicles' model/sensor stacks are proprietary, sharing parameters or retraining on private data is often not feasible. The second approach introduces feature interpreters that translate or align heterogeneous features to the unified feature space, preserving single-agent performance and model confidentiality~\cite{b13,b14,b15}. However, these interpreters usually rely on larger, more expensive models and still need retraining or extra storage whenever a new heterogeneous agent joins. These limitations motivate methods that generalize efficiently to unseen agents while maintaining accuracy, privacy, and computational feasibility.

%In this paper, we propose \textsc{Faster-HEAL}, a heterogeneous collaborative perception framework that efficiently bridges the semantic domain gap between heterogeneous agents while preserving privacy and maintaining significantly higher computational efficiency relative to the original HEAL method. Our approach, which deploys the initial stage of HEAL \cite{b12}, introduces a novel Lightweight Interpreter for Feature Transformation (LIFT). The proposed LIFT strategy aligns intermediate features from newly joining heterogeneous neighbor agents with the unified feature space of the ego agent. Hence, the framework is trained in two stages: first, the ego agent is trained on a homogeneous collaborative perception base to construct a unified feature space and fusion model; second, the lightweight interpreter is trained to map new heterogeneous agents' features into the unified feature space without retraining any other model components. This design preserves the privacy and single-agent accuracy of newly added agents, avoids costly retraining and storage overhead, and enables scalable deployment. By leveraging a unified feature space together with the proposed parameter-efficient LIFT interpreter, \textsc{Faster-HEAL} mitigates the performance degradation typically observed in heterogeneous collaborative settings while maintaining computational efficiency.
In this paper, we propose \textbf{Faster-HEAL}, a heterogeneous collaborative perception framework that efficiently bridges the semantic domain gap between heterogeneous agents while preserving privacy and improving computational efficiency over HEAL~\cite{b12}. Building on the first stage of HEAL, we introduce a novel \textbf{L}ightweight \textbf{I}nterpreter for \textbf{F}eature \textbf{T}ransformation (\textbf{LIFT}) that aligns intermediate features from new heterogeneous neighbor agents with the ego agent’s unified feature space. The framework is trained in two stages: first, the ego agent is trained on a homogeneous CP base to construct the unified feature space and fusion model; second, \textsc{LIFT} is fine-tuned to map new heterogeneous agents’ features into that space along with two other small components. This design preserves the privacy and single-agent accuracy of new agents, avoids costly retraining and storage overhead, and enables scalable deployment. By combining the unified feature space with \textsc{LIFT}, \textsc{Faster-HEAL} mitigates performance degradation in heterogeneous settings while maintaining high computational efficiency, and preserving privacy.

%We leverage visual prompts as the core mechanism for our \textsc{LIFT}-based feature interpretation. Analogous to language prompts, visual prompts guide the model by emphasizing task-relevant information in the feature representation.Specifically, we reuse the single-agent pretrained encoder and backbone of the new agent, along with the pretrained pyramid fusion module and detection head of the ego agent, and train only the vision prompt parameters to align heterogeneous features. By freezing all pretrained components, our approach preserves single-agent detection accuracy and protects the privacy of newly introduced agents.Although naively training visual prompts can be computationally expensive due to their high dimensionality, we design a low-rank prompt representation that reduces the number of trainable parameters by an order of magnitude, significantly improving training efficiency.
We leverage visual prompts as the core mechanism in our \textsc{LIFT}-based feature interpreter, which guides the model by emphasizing task-relevant information in the feature representation, like language prompts. Since visual prompts have high dimensionality, naively training them is computationally expensive. We therefore design a low-rank prompt representation that reduces the number of trainable parameters by an order of magnitude and significantly improves training efficiency.

%Our key contributions are summarized as follows:
%\begin{itemize}
   %\item We design \textbf{Faster-HEAL}, a lightweight feature interpreter (LIFT) for newly emerging heterogeneous agent types, enabling single-stage feature alignment with the ego agent’s unified feature space.  Only a visual prompt is fine-tuned for each new agent type with a different modality. Hence, the ego agent only requires intermediate features from collaborators, allowing each vehicle to keep its sensor configuration and model parameters private. 
   % \item By introducing low-rank visual prompts, our approach achieves fast and accurate feature adaptation while minimizing the number of trainable parameters. Moreover, our method reduces computation cost by an order of magnitude and requires the ego agent to store only the visual prompts corresponding to different agent types.
    %\item Experimental results show that \textsc{Faster-HEAL} improves state-of-the-art detection performance by at least $2\%$ while significantly reducing computational overhead by at least $90\%$, representing a practical step toward scalable and privacy-preserving collaborative perception in real-world autonomous driving.
%\end{itemize}
Our key contributions are summarized as follows:
\begin{itemize}
\item We design \textbf{Faster-HEAL}, which uses a lightweight interpreter for feature translation \textbf{(LIFT)} for newly emerging heterogeneous agent types, enabling single-stage feature alignment with the ego agent’s unified feature space. For each new agent type with a different modality, only a visual prompt is fine-tuned along with a feature aligner, so the ego agent needs only intermediate features, and each vehicle keeps sensor/model configurations private.
\item By introducing low-rank visual prompts as \textsc{LIFT}, our approach achieves fast and accurate feature adaptation while reducing trainable parameters and computation by an order of magnitude, and lowering storage overhead, since only the \textsc{LIFT} for different agent types are stored.
\item Experimental results show that \textsc{Faster-HEAL} improves state-of-the-art detection performance by  $2\%$ while reducing computational overhead by $94\%$ on average for training, representing a practical step toward scalable and privacy-preserving CP in real-world autonomous driving.
\end{itemize}

\section{Related Work}
\subsection{Collaborative Perception}
%Sharing perception data among connected agents through collaborative perception significantly improves the accuracy and reliability of downstream perception tasks, particularly in occluded or long-range scenarios. Collaborative perception is typically categorized into three levels based on the granularity of shared information \cite{b1}. Early fusion transmits raw sensor data among agents, achieving maximal perception performance but incurring substantial communication bandwidth requirements. Late fusion, in contrast, exchanges only final perception results, reducing bandwidth consumption but suffering from performance degradation due to noise accumulation and inconsistent predictions. To balance these trade-offs, recent work has focused on intermediate fusion, which shares processed feature representations to jointly optimize communication efficiency and perception quality. For instance, Where2com \cite{b2} and When2com \cite{b5} dynamically select communication partners and timing to alleviate bandwidth constraints while preserving performance, with Codefilling\cite{b6} pursuing a similar objective. Other studies have addressed challenges such as noisy communication and latency, as well as pose estimation errors that misalign shared features \cite{b3}, \cite{b4},\cite{b7}. Nevertheless, most existing frameworks assume homogeneous sensors and models across agents, leaving the problem of heterogeneity an important open direction for future research.
Sharing perception data among connected agents through collaborative perception improves the accuracy and reliability of downstream tasks, especially in occluded or long-range scenarios. CP is usually categorized into three levels based on the granularity of shared and fused information~\cite{b1}: (a) early fusion, which transmits raw sensor data and can maximize perception performance but requires very high bandwidth; (b) late fusion, which exchanges only final detection results, saving bandwidth but suffering from noise accumulation and inconsistent predictions; (c) intermediate fusion, which shares processed feature representations to balance communication efficiency and perception quality. Recent intermediate-fusion methods such as Where2com~\cite{b2} and Codefilling~\cite{b6} dynamically select communication partners, timing, or content to relieve bandwidth pressure while preserving performance. Other studies further tackle noisy communication, latency, and pose estimation errors that misalign shared features~\cite{b3,b4,b7}. However, most existing frameworks still assume homogeneous sensors and perception models across agents, leaving heterogeneity as a key open challenge.
\subsection{Heterogeneous Collaborative Perception}
Addressing the performance degradation caused by the feature domain gap among heterogeneous agents has become an active research area. Most approaches mitigate this gap by retraining the whole network or part of it to align features across agents. DI-V2X~\cite{b8}, HM-ViT~\cite{b9}, DGCNN~\cite{b10}, and CoCMT~\cite{b11} each address the domain gap for each agent type by retraining the model or selected submodules, which increases computational overhead, violates neighbors' privacy, and adapts poorly to new agent types at deployment. HEAL~\cite{b12} further proposes a pyramid fusion model that builds a unified feature space but still requires retraining the new agent’s encoder, again raising data privacy concerns. To avoid modifying full perception stacks, MDPA~\cite{b13} and PnPDA+~\cite{b14} use feature interpreters to align new-agent features to a unified feature space, reducing the amount of retraining while keeping the original perception models intact. Specifically, PolyInter~\cite{b15} adopts visual prompts alongside attention modules as interpreters. These approaches limit retraining to a dedicated module, but they often incur higher model complexity and computational overhead.

\subsection{Visual Prompts}
%Prompt-based learning has recently emerged as a powerful tool for domain adaptation in vision tasks, enabling models to bridge feature distribution gaps with minimal parameter updates. By injecting learnable visual tokens into the feature space, visual prompts guide the network to adapt to new domains and improve downstream task performance without full model retraining. AD-CLIP~\cite{b16}, ViP-LLaVA~\cite{b17}, and UPR~\cite{b19} used visual prompts to align the domain gap for domain classification, using domain-invariant features. On the other hand, DPR~\cite{b18} and ModPrompt~\cite{b21} addressed the detection performance degradation caused by heterogeneous features, using visual prompt tuning. Collectively, these works establish visual prompts as a compact yet expressive mechanism for domain alignment, offering strong performance gains with minimal computational cost.
Prompt-based learning has recently emerged as a powerful tool for domain adaptation in vision tasks, enabling models to bridge feature distribution gaps with minimal parameter updates. By injecting learnable visual tokens into the feature space, visual prompts guide the network to adapt to new domains and improve downstream task performance without full model retraining. AD-CLIP~\cite{b16}, ViP-LLaVA~\cite{b17}, and UPRE~\cite{b19} use visual prompts to mitigate domain gaps for classification by learning domain-invariant features. In contrast, DPR~\cite{b18} and ModPrompt~\cite{b21} address detection performance degradation under heterogeneous features via visual prompt tuning. Collectively, these works establish visual prompts as a compact mechanism for domain alignment, offering strong performance gains with minimal computation.

\section{Methodology}
%The overall architecture of \textsc{Faster-HEAL} is illustrated in Figure \ref{fig:placeholder}. Our framework consists of two stages designed to address the challenge of feature misalignment in heterogeneous collaborative perception. In the first stage, we adopt the homogeneous base training procedure from HEAL~\cite{b12}, using the same model architecture and loss formulation to construct a unified feature space with the ego agent. A brief summary appears in Subsection~\ref{subsec:homogeneous-base} for completeness. In the second stage, we fine-tune lightweight visual prompts to align the intermediate features of a newly introduced heterogeneous agent with the unified feature space of the ego agent. For each new agent type, only the visual prompts need to be tuned, enabling rapid adaptation while preserving both privacy and single-agent performance.

\begin{figure*}
    \centering
\includegraphics[width=0.70\linewidth]{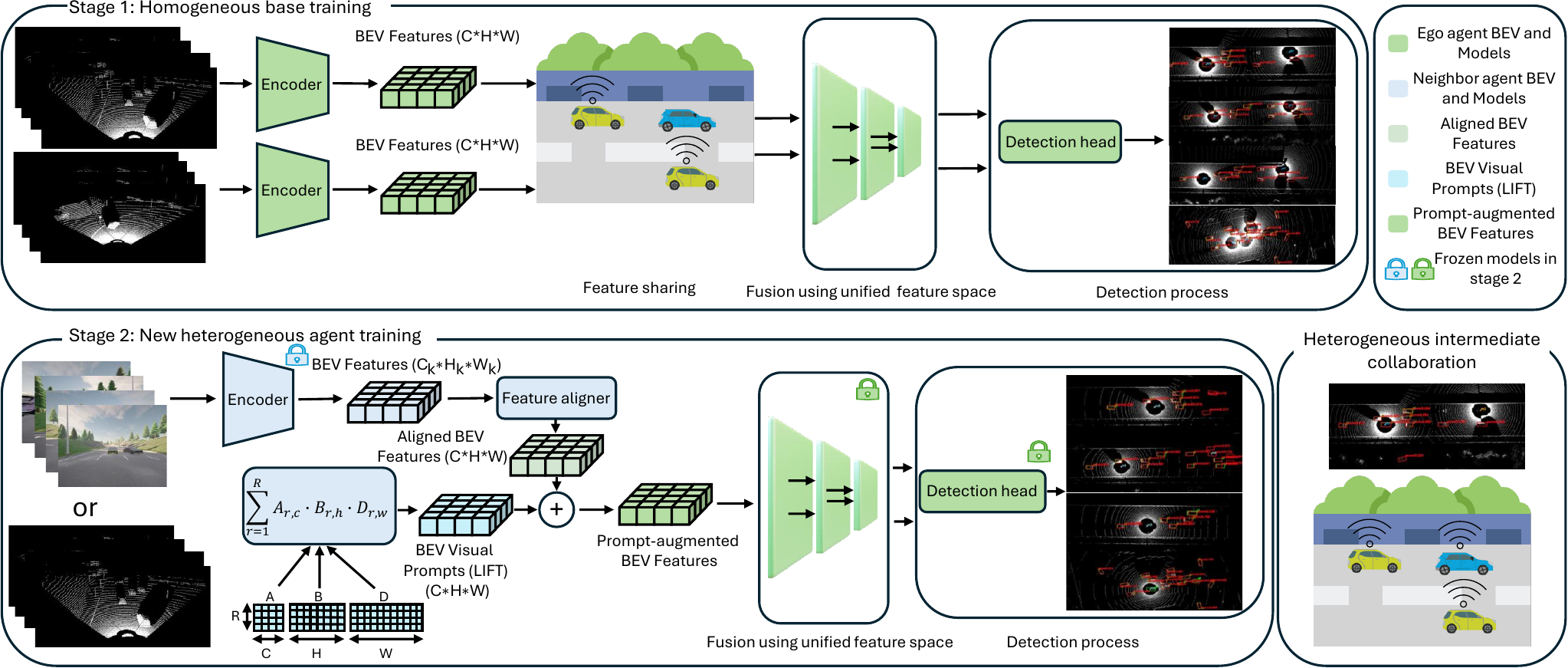}
    \caption{Overview of \textbf{Faster-HEAL}. Stage~1: Homogeneous base training constructs a unified feature space using pyramid fusion and trains the detection head on collaborative data from homogeneous agents. stage~2: For each new heterogeneous agent type, we freeze the pretrained encoder, pyramid fusion extractor/upsampling path, and detection head, and train a lightweight feature aligner, \textsc{LIFT} (visual prompts) to map its intermediate features into the unified space, and the foreground estimator. This design ensures low training cost, preserves the privacy of participating agents by requiring only their intermediate features, and enables scalable heterogeneous collaboration. In our experiments, LiDAR is used in stage~1, while camera or LiDAR could be used in stage~2.}

    \label{fig:overall}
\end{figure*}

The overall architecture of \textsc{Faster-HEAL} is shown in Fig.~\ref{fig:overall}. The framework has two stages to address feature misalignment in heterogeneous collaborative perception. In stage~1, we adopt the homogeneous base training architecture and loss of HEAL~\cite{b12}, to learn a unified feature space for the ego agent. A brief summary appears in Subsection~\ref{subsec:homogeneous-base}. In stage~2, we introduce and fine-tune \textbf{LIFT} as lightweight visual prompts to align the newly introduced heterogeneous intermediate features with the ego agent unified space, enabling rapid adaptation (to the best of our knowledge, among the lowest interpreter-based methods) while preserving privacy and single-agent performance (unlike retraining methods).

%Fig.~\ref{fig:overall} overviews \textsc{Faster-HEAL}. It has two stages to address feature misalignment in heterogeneous collaborative perception. In stage~1, we adopt the homogeneous base training architecture and loss of HEAL~\cite{b12} to learn a unified feature space for the ego agent (Subsection~\ref{subsec:homogeneous-base}). In stage~2, we introduce and fine-tune \textbf{LIFT} prompts to align new heterogeneous intermediate features to the ego unified space, enabling rapid adaptation with \emph{minimal} per-type trainable overhead (to the best of our knowledge, the lowest among interpreter-based methods) while preserving privacy and single-agent performance.

\subsection{Homogeneous Base Training}
\label{subsec:homogeneous-base}
%In the first stage of \textsc{Faster-HEAL}, we train a unified feature space using $N$ homogeneous neighbor agents together with the ego agent. For each agent $i \in \{1,2,\dots,N\}$, the observation $O_i$ is passed through the encoder $f_{enc\_ego}$ to produce BEV features $F_i$, which are then spatially transformed and transmitted to the ego agent as $F_{i \rightarrow e}$ via the transformation and communication function $\Gamma_{i \rightarrow e}$. Note that $e \in [1,N]$ and $F_{e \rightarrow e} = F_e$ is the ego features that do not need transformation and communication. All received features $\{F_{i \rightarrow e}\}$ are fused with $F_e$ using the pyramid fusion module \cite{b12} $f_{\text{pyramid}}$ to produce the unified feature representation $H_e$, which is then passed through the detection head $f_{\text{head}}$ to generate the final detection results $B$:
In the first stage of \textsc{Faster-HEAL}, we learn a unified feature space using the ego agent (e) and $N-1$ homogeneous neighbors. For each agent $i \in \{1,2,\dots,N-1,N\} (N\: \text{is for e)}$, the observation $O_i$ is encoded by the shared ego encoder $f_{enc\_ego}$ into BEV features $F_i$, which are transformed to the ego frame and transmitted as $F_{i \rightarrow e}$ via communication function $\Gamma_{i \rightarrow e}$. The ego then fuses all received features $\{F_{i \rightarrow e}\}$ using the pyramid fusion module $f_{\text{pyramid}}$ to obtain the ego agent unified fused representation $H_e$, which is passed to the detection head $f_{\text{head}}$ to produce the final detections $B$:
\begin{equation}
    \begin{gathered}
        F_i = f_{enc\_ego}(O_i),  i = 1,2,\dots,N \\
        F_{i \rightarrow e} = \Gamma_{i \rightarrow e}(F_i),  i = 1,2,\dots,N \\
        H_e = f_{\text{pyramid}}(F_{1 \rightarrow e}, F_{2 \rightarrow e}, \dots, F_{N \rightarrow e}) \\
        B = f_{\text{head}}(H_e)
    \end{gathered}
\end{equation}
%Here, $\Gamma_{i \rightarrow e}$ denotes the spatial transformation and communication process, $F_{i \rightarrow e}$ represents the feature transmitted from agent $i$ to the ego agent, $F_e$ corresponds to the ego agent’s BEV features, $H_e$ is the fused feature representation obtained through pyramid fusion, and $B$ represents the final detection results.%

%To enable efficient and accurate feature fusion, we adopt a Pyramid Fusion \cite{b12} module. A ResNeXt \cite{b22} network is employed to progressively downsample the incoming features by a factor of two, yielding $L$ multi-scale feature maps. At each scale, a foreground estimator generates an occupancy map, which is normalized via a softmax function to compute per-agent fusion weights. The weighted features are then aggregated, upsampled to a common resolution, and concatenated to produce the final fused representation $H_e$ for the detection head:
To enable efficient and accurate feature fusion, we adopt the Pyramid Fusion module from~\cite{b12}. A ResNeXt network~\cite{b22} progressively downsamples incoming features by factor of two, producing $L$ multi-scale features. At each scale $l$, a foreground estimator generates an occupancy map that is normalized across agents to obtain fusion weights. The weighted features are then aggregated, upsampled to a common resolution, and concatenated to form fused features $H_e$:
\begin{equation}
    \begin{gathered}
        F_{i \rightarrow e}^{(l)} = f^{(l)}_{\text{ResNeXt}}\!\left(F_{i \rightarrow e}^{(l-1)}\right),  i = 1,2,\dots,N,\; l=1,\dots,L \\
        OCC_{i \rightarrow e}^{(l)} = f_{\text{foreground}}^{(l)}\!\left(F_{i \rightarrow e}^{(l)}\right),  i = 1,2,\dots,N,\; l=1,\dots,L \\
        W_{i \rightarrow e}^{(l)} = \text{softmax}\!\left(OCC_{1 \rightarrow e}^{(l)},\dots,OCC_{N \rightarrow e}^{(l)}\right), \\
        F_e^{(l)} = \sum_{i=1}^{N} F_{i \rightarrow e}^{(l)} \cdot W_{i \rightarrow e}^{(l)}, \\
        F_e^{(l)} = f^{(l)}_{\text{upsample}}\!\left(F_e^{(l)}\right),  l = 1,\dots,L \\
        H_e = \text{concat}\!\left(F_e^{(1)},F_e^{(2)},\dots,F_e^{(L)}\right),
    \end{gathered}
\end{equation}
%Here, $l$ denotes the scale level, $F_{i \rightarrow e}^{(l)}$ represents the encoded features of agent $i$ at scale $l$, and $f_*^{(l)}$ indicates the corresponding operation (e.g., ResNeXt downsampling or upsampling) at scale $l$.
where $l$ denotes the scale level, and $f_{*}^{(l)}$ indicates the corresponding ResNeXt, upsampling, or foreground estimation.

%Using multi-scale features in the unified feature space leverages both coarse and fine-grained representations, enhancing the adaptivity of the system when fusing BEV features from multiple agents. Higher scales further improve robustness by reducing the impact of partial transformation errors. Moreover, the foreground estimator focuses on the most informative regions of each agent’s feature map, leading to improved fusion quality and downstream detection performance.
Using multi-scale features in the unified feature space leverages coarse-to-fine features, improving fusion adaptivity and robustness to transformation errors, while the foreground estimator focuses on informative regions of each agent’s feature map to enhance downstream detection.

%The total training loss for the homogeneous base stage is defined as the sum of the final detection loss and the supervised foreground estimation loss computed at each scale level:
The total training loss for the homogeneous base stage is the sum of the final detection loss and the supervised foreground estimation losses at all scales:
\begin{equation}
    \begin{gathered}
        L = L_{\text{det}}(B,Y) + \sum_{l=1}^{L} \alpha_l \, L_{\text{foreground}}^{(l)} \\
        L_{\text{det}}(B,Y) = L_{\text{focal}}(B,Y)\\ 
        + 2 \cdot L_{\text{smooth-L1}}(B,Y)
        + 0.2 \cdot L_{\text{dir}}(B,Y) \\
        L_{\text{foreground}}^{(l)} = \sum_{i=1}^{N} L_{\text{focal}}\!\left(OCC_{i \rightarrow e}^{(l)},Y_{i \rightarrow e}^{(l)}\right),
    \end{gathered}
\end{equation}
%Here, $L_{\text{focal}}$ is the focal loss \cite{b23} used for classification, $L_{\text{smooth-L1}}$ \cite{b24} is the regression loss, and $L_{\text{dir}}$ is the direction regression loss. $Y$ denotes the ground-truth detections and $B$ the predicted detections. At each scale $l$, a supervised foreground estimator is trained using the focal loss, where $OCC_{i \rightarrow e}^{(l)}$ is the predicted occupancy (foreground) map for agent $i$ and $Y_{i \rightarrow e}^{(l)}$ is the corresponding ground-truth BEV mask. The hyperparameter $\alpha_l$ controls the relative contribution of the foreground estimator loss at each scale.
where, $L_{\text{focal}}$, $L_{\text{smooth-L1}}$, and $L_{\text{dir}}$ denote the focal classification loss, regression loss, and direction loss, respectively; $Y$ and $B$ are the ground-truth and predicted detections. At each scale $l$, the foreground estimator is supervised by focal loss on the predicted occupancy map and its BEV mask $Y_{i \rightarrow e}^{(l)}$, with $\alpha_l$ weighting the contribution of each scale.
\subsection{New heterogeneous agent training}
%After training the homogeneous base, a new heterogeneous agent may be introduced into the collaboration group. However, the BEV feature domain gap—caused by differences in sensors, perception models, or even sensor configurations—can significantly degrade collaborative performance compared to the homogeneous case. As a result, the ego agent cannot directly use the raw intermediate features from the new agent and must first transform them to align with its unified feature space. In real-world deployments, however, computational and time resources are limited, and retraining the new agent’s model is typically infeasible. Moreover, manufacturers are often unwilling to share model parameters or modify proprietary networks, making full retraining impractical. These challenges motivate the need for a lightweight, privacy-preserving mechanism to adapt heterogeneous agents efficiently.
After training the homogeneous base, new heterogeneous agents may join the collaboration group. The BEV feature domain gap caused by different sensors, perception models, or sensor configurations can significantly degrade performance, so the ego agent cannot directly use their raw intermediate features and must first align them with its unified feature space. In real-world deployments, computational and time resources are limited, and retraining the new agent’s model is often infeasible, while manufacturers are typically unwilling to share parameters or modify proprietary networks. These constraints motivate a lightweight, privacy-preserving mechanism to adapt heterogeneous features efficiently.

%The key idea of \textsc{Faster-HEAL} in this stage is to adapt the features of newly introduced heterogeneous agents while preserving the unified feature space of the ego agent. To achieve this, the fusion and detection modules of the ego agent and the encoder of the new agent are kept frozen, ensuring that the learned feature representation remains consistent. This design also preserves the privacy of the new agents, since only their intermediate features are shared, without requiring any access to their sensor configurations or model parameters.
The key idea of \textsc{Faster-HEAL} in this stage is to adapt features from newly introduced heterogeneous agents while preserving the ego agent’s unified feature space. We freeze the ego fusion extractor/upsampling and detection modules as well as the new agent’s encoder, so the learned representation stays consistent, and only intermediate features are shared, which preserves the new agent’s privacy by avoiding access to its sensor configurations or model parameters, unlike HEAL~\cite{b12}.

%To close the domain gap, \textsc{Faster-HEAL} injects learnable visual prompts into the transmitted intermediate features. These prompts are fine-tuned for each new agent type, producing a representation that is compatible with the ego agent’s feature space. For a new agent $k$ of type $n_1$ with observation $O_k$, and with its encoder $f_{enc\_{n_1}}$ kept frozen, the training procedure in the second stage is formulated as:
To close the domain gap, \textsc{Faster-HEAL} injects learnable visual prompts as \textsc{LIFT} to the transmitted intermediate features, fine-tuning a separate prompt for each new agent type to make its representation compatible with the ego agent’s unified feature space. For a new agent $k$ of type $n_1$ with observation $O_k$ and frozen encoder $f_{enc\_{n_1}}$, the second-stage training is:
\begin{equation}
    \begin{gathered}
        F_k = f^{**}_{enc\_{n_1}}(O_k),\\
        F_{k \rightarrow e} = \Gamma_{k \rightarrow e}(F_k), \\
        F'_k = f_{\text{aligner}_{n_1}}(F_{k \rightarrow e})\textbf{},\\ 
        F''_k = F'_k + P_{n_1}, \\
        H_{e,k} = f^{*}_{\text{pyramid}}(F''_k), \\
        B_k = f^{*}_{\text{head}}(H_{e,k}),
    \end{gathered}
\end{equation}
%where, $F_k$ is the intermediate BEV feature of agent $k$, $F_{k \rightarrow e}$ is the transmitted feature after spatial transformation, and $F'_k$ is the channel-aligned feature obtained using the aligner $f_{\text{aligner}_{n_1}}$, which employs a ConvNeXt-based \cite{b25} projection to match the feature dimensions of the ego agent. $P_{n_1}$ denotes the learnable visual prompt for agent type $n_1$, and $F''_k$ is the prompt-augmented feature provided as input to the frozen pyramid fusion module. $H_{e,k}$ represents the feature of agent k in the ego unified representation, and $B_k$ is the final detection result. The superscript $*$ indicates that the corresponding module is pretrained and remains frozen during this stage.
where $F_k$ is the intermediate BEV feature of agent $k$, $F_{k \rightarrow e}$ is its spatially transformed feature, and $F'_k$ is the channel-aligned output of the aligner $f_{\text{aligner}_{n_1}}$, which employs a ConvNeXt-based~\cite{b25} projection that maps ($C_k, H_k,W_k$) to ($C, H,W$). $P_{n_1}$ is the learnable visual prompt for agent type $n_1$, $F''_k$ is the prompt-augmented feature fed to the frozen pyramid fusion module, $H_{e,k}$ is agent $k$’s features in the ego unified representation, and $B_k$ is the final detection. The superscript $*$ (from stage~1), $**$ (from neighbor) indicates the corresponding module is pretrained and kept frozen in this stage.

%To ensure compatibility, the prompt $P_{n_1}$ must have the same number of channels, height, and width $(C,H,W)$ as the aligned features of the neighbor, which match the features of the ego agent. The prompt is randomly initialized and fine-tuned using the ground-truth detection loss. However, even a small BEV feature map contains $C \times H \times W$ parameters, which can exceed the parameter count of a typical encoder, leading to high computational cost and slow convergence.
%To address this challenge, \textsc{Faster-HEAL} applies PARAFAC (CP) decomposition \cite{b36} to factorize the visual prompt into a low-rank representation. Let $R$ denote the decomposition rank. Three learnable 2D tensors are initialized: $A \in \mathbb{R}^{R \times C}$, $B \in \mathbb{R}^{R \times H}$, and $D \in \mathbb{R}^{R \times W}$, where $C$, $H$, and $W$ denote the channel, height, and width dimensions, respectively. The prompt is then reconstructed as
To ensure compatibility, the visual prompt $P_{n_1}$ has the same channels, height, and width $(C,H,W)$ as the aligned neighbor features, which match the ego features. $P_{n_1}$  is randomly initialized and fine-tuned using the ground-truth detection loss. However, even a small BEV prompt contains $C \times H \times W$ trainable parameters, which may exceed the size of a typical encoder and lead to high computational cost and slow convergence. To address this challenge, \textsc{Faster-HEAL} applies PARAFAC decomposition~\cite{b36} to factorize the prompt into a low-rank representation.  Let $R$ denote the decomposition rank; three 2D tensors are initialized as $A \in \mathbb{R}^{R \times C}$, $B \in \mathbb{R}^{R \times H}$, and $D \in \mathbb{R}^{R \times W}$, where $C$, $H$, and $W$ are the channel, and spatial dimensions. The prompt is then reconstructed as
\begin{equation}
\label{equ:cp}
    P_{c,h,w} \approx \sum_{r=1}^{R} A_{r,c} \cdot B_{r,h} \cdot D_{r,w},
\end{equation}
%where $P_{c,h,w}$ represents the prompt element at channel $c$ and spatial position $(h,w)$. This factorization reduces the number of trainable parameters from $C \times H \times W$ to $R \times (C + H + W)$, yielding several orders of magnitude fewer parameters (from millions to only a few thousand). As a result, training costs are significantly reduced, convergence is faster, and experimental results show that fine-tuning the decomposed prompt achieves performance comparable to full-prompt tuning at a fraction of the computational overhead.
where $P_{c,h,w}$ is the prompt element at channel $c$ and spatial location $(h,w)$. This factorization reduces the trainable parameters from $C \times H \times W$ to $R \times (C + H + W)$ with $R \ll min(C,H,W)$, shrinking them from millions to a few thousand, which lowers training cost, speeds convergence, and, retains performance comparable to full-prompt tuning at a fraction of the computational overhead.
%Training the aligner, foreground estimator, and visual prompts jointly ensures effective feature alignment and robust multi-scale domain adaptation. The aligner, implemented using a ConvNeXt-based channel projection, matches the feature dimensions between agents, while the learnable visual prompts inject task-relevant information to reduce the domain gap. The foreground estimator is retrained to emphasize domain-specific salient regions, further improving fusion quality.
%By freezing the encoder of the new agents, \textsc{Faster-HEAL} preserves privacy by eliminating the need to share model weights or sensor parameters, while the centralized training of the ego agent removes the requirement for retraining on other agents. This design also reduces storage requirements, as only the small prompt parameters need to be stored rather than full encoder models for every agent type. Once stage 2 is completed for heterogeneous types $n_1$, $n_2$, $\dots$, the ego agent can collaborate effectively with any agent of the same heterogeneous types $n_1$, $n_2$, $\dots$, and the prompt-based adaptation can be retrained as needed, enhancing the robustness and scalability of the collaborative perception framework.

%Training the aligner, foreground estimator, and \textsc{LIFT} (visual prompts) jointly ensures effective feature alignment and robust multi-scale domain adaptation, 
%The aligner, implemented as a ConvNeXt-based channel projection, matches feature dimensions across agents, while the learnable visual prompts inject task-relevant information to reduce the domain gap, and 
Training the aligner primarily resolves dimensional/statistical mismatch, while \textsc{LIFT} (visual prompts) ensures effective feature alignment and robust multi-scale domain adaptation, 
and retraining the foreground estimator for new agents emphasizes domain-specific salient regions and improves fusion quality. By freezing the encoders of new agents, \textsc{Faster-HEAL} preserves privacy by avoiding any sharing of model weights or sensor parameters, and centralized training at the ego side removes the need to retrain other agents. This design also reduces storage, since only compact \textsc{LIFT} (prompt parameters) are stored instead of full encoder models for each agent type. Once stage~2 is completed for heterogeneous types $n_1, n_2, \dots$, the ego agent can collaborate with any agent of these types, and the prompt-based adaptation can be further retrained if needed, enhancing the robustness and scalability of the collaborative perception framework. The loss function used in this stage is identical to that of the first training stage, combining detection loss and supervised foreground losses across scales.

\subsection{Inference Implementation}
%After completing stage-two training for a heterogeneous neighbor agent, the ego agent shares the trained aligner with the heterogeneous neighbor to enable collaborative perception. The aligner allows the neighbor’s intermediate features to be compressed before transmission, minimizing the required communication bandwidth. Additionally, the privacy of all agents is preserved since none of them share their complete models or sensor specifications. The ego agent selects the appropriate visual prompt based on the neighbor’s sensor and encoder type, without requiring access to proprietary model parameters.
%When a new agent joins the collaboration at deployment, the ego agent shares the corresponding aligner and chooses the appropriate prompt for detection based on the agent’s sensor and model type. If the model parameters differ from those encountered during stage-two training, a slight performance degradation may occur; however, the system remains functional. If desired, the ego agent can fine-tune the prompts for a few epochs to adapt to the new agent. Thanks to the domain similarity between models of the same type and the low number of trainable parameters (fewer than 10k), this adaptation can be performed rapidly, approaching real-time performance.
After completing stage~2 training for a heterogeneous neighbor, the ego agent shares the trained aligner with that neighbor to enable collaborative perception. The aligner compresses the neighbor’s intermediate features before transmission, reducing communication bandwidth, while privacy is preserved since no agent shares its full model or sensor specifications. At deployment, 
%when a new agent joins, the ego agent selects the corresponding \textsc{LIFT} (visual prompt) based on the agent’s sensor and encoder type-ID 
each agent includes a type-ID (e.g., sensor and encoder modality). Then the ego uses this ID to select the matching \{aligner, LIFT\} pair, without needing access to proprietary sensor or model parameters, and shares the matching aligner. 
 If the new agent’s model parameters differ from those seen during stage~2, a slight performance degradation may occur, but the system remains functional. If desired, the ego can fine-tune the \textsc{LIFT} for a few epochs to better adapt to this agent. Due to the domain similarity between models of the same type and the very small number of trainable prompt parameters (fewer than 10k), this adaptation can be performed rapidly, approaching real-time performance.

\section{Experiments}
\subsection{Datasets}
We evaluate \textsc{Faster-HEAL} on OPV2V-H dataset to assess its performance. OPV2V-H~\cite{b12} is a public dataset built upon OPV2V~\cite{b26}, a large-scale benchmark for collaborative perception in autonomous driving. 
%While OPV2V provides a diverse set of scenarios, LiDAR exhibits a strong detection advantage over cameras in its original configuration. To mitigate this imbalance, OPV2V-H collected additional camera data to ensure a more balanced representation between LiDAR and camera modalities. The dataset contains scenes with at least two and up to seven collaborating agents, with an average of three agents per scenario. Each agent is equipped with 16-, 32-, or 64-channel LiDAR sensors and four RGB cameras with depth information.

%In addition, we utilize DAIR-V2X, a real-world dataset designed for vehicle–infrastructure cooperative perception. It consists of one vehicle equipped with a 40-channel LiDAR and a camera, and one infrastructure node equipped with a 300-channel LiDAR and a camera, providing a realistic setting for evaluating heterogeneous sensor configurations.

\subsection{Scenario Design}
%To evaluate the generalization capability of \textsc{Faster-HEAL} across different sensor types and perception encoders, we design scenarios involving both LiDAR- and camera-based agents. For LiDAR, we adopt \textit{PointPillar} \cite{b27} and \textit{SECOND} \cite{b28} as feature encoders, which are widely used in collaborative perception research. For camera-based agents, we use \textit{EfficientNet} \cite{b29} and \textit{ResNet-50} \cite{b30} as image encoders in conjunction with \textit{Lift-Splat-Shoot} \cite{b31}for BEV feature generation.
To evaluate the generalization capability of \textsc{Faster-HEAL} across different sensor types and perception encoders, we design scenarios with both LiDAR- and camera-based agents. For LiDAR, we adopt \textit{PointPillar}~\cite{b27} and \textit{SECOND}~\cite{b28} as feature encoders. For camera-based agents, we use \textit{EfficientNet}~\cite{b29} and \textit{ResNet-50}~\cite{b30} as image encoders in conjunction with \textit{Lift-Splat-Shoot}~\cite{b31} for BEV feature generation.

To ensure comparability with prior work, we follow the HEAL~\cite{b12} framework and train the homogeneous base using a 64-channel LiDAR agent ($m_1$) with PointPillar to construct the unified feature space. For heterogeneous experiments, we add three new agent types: (i) a camera-based agent with EfficientNet ($m_2$), (ii) a LiDAR agent using SECOND ($m_3$), and (iii) a camera-based agent with ResNet-50 ($m_4$). All LiDAR and camera encoders use a BEV grid resolution of $[0.4\,\mathrm{m}, 0.4\,\mathrm{m}]$, and their encoded features pass through a BEV backbone before fusion. The aligner uses a ConvNeXt~\cite{b25} projection to match feature size to the ego space, and the pyramid fusion module operates over three scales 
with $[64, 128, 256]$ 
and $[3, 5, 8]$ ResNeXt~\cite{b22} blocks, 
and $1 \times 1$ foreground estimator at each scale. All new agents use pretrained and frozen single-agent encoders and BEV backbones to avoid retraining large models. Both the base training stages run for 25 epochs with Adam (learning rate $0.002$). 
The detection range is $[-102.4\,\mathrm{m}, +102.4\,\mathrm{m}]$ along $x$ and $[-51.2\,\mathrm{m}, +51.2\,\mathrm{m}]$ along $y$. We evaluate average precision (AP) at different intersection-over-union (IoU) thresholds and, during testing, initialize each scenario with the base agent and then progressively introduce new heterogeneous agent types to assess adaptability.
\subsection{Results}
We compare \textsc{Faster-HEAL} with state-of-the-art intermediate fusion frameworks for heterogeneous collaborative perception. As summarized in Table~\ref{tab:per}, \textsc{Faster-HEAL} improves HEAL~\cite{b12} by about $1.8\%/1.3\%$ at AP@0.5/0.7 IoU and outperforms CoBEVT~\cite{b35} by $5.3\%/12\%$ at AP@0.5/0.7, showing that fine-tuning \textsc{LIFT} is an effective way to handle sensor and model heterogeneity of new agents without retraining the whole or a part of the perception stacks. 
%These gains mainly stem from the multi-scale pyramid fusion, which adaptively weights features by their contribution to detection, and from the prompts, which close the domain gap for new agents. 
The last three columns of Table~\ref{tab:per} report the number of trainable parameters where \textsc{Faster-HEAL} reduces them from tens of millions (as in HEAL and other retraining-based approaches) to only a few hundred thousand, a reduction of roughly $90\%$, while improving accuracy, makes \textsc{Faster-HEAL} practical for real-world deployments, where fast adaptation and low storage overhead are crucial. Training efficiency in Table~\ref{tab:fast} further shows that \textsc{Faster-HEAL} reaches $16.64$ TFLOPs/s, about $2.35\times$ higher hardware utilization and $1.2\times$ higher end-to-end training throughput than HEAL, while lowering peak GPU memory by $38\%$ on an NVIDIA RTX A6000 with batch size 1, confirming \textsc{Faster-HEAL} is both accurate and computationally efficient.

\begin{table*}[!t]
\centering
\caption{Detection performance comparison on OPV2V-H using $m_1$ as the collaborative base and progressively adding three new heterogeneous agents ($m_2$, $m_3$, $m_4$). We report AP@0.5, AP@0.7, and the model number of trainable parameters for No-Fusion, Late Fusion (no retraining), HEAL~\cite{b12} (encoder retraining), and CoBEVT~\cite{b35} that retrain their entire model.}
%state-of-the-art (SOTA) methods that retrain their entire model. \textsc{Faster-HEAL} achieves comparable or superior performance with significantly fewer trainable parameters.}
\label{tab:per}

\resizebox{0.9\textwidth}{!}{
\begin{tabular}{lccccccccc}
\toprule
\textbf{Metric} & \multicolumn{3}{c}{\textbf{AP@0.5(\%)} $\uparrow$} & \multicolumn{3}{c}{\textbf{AP@0.7(\%)} $\uparrow$} & \multicolumn{3}{c}{\textbf{Model \#Training Params (M)} $\downarrow$} \\
\cmidrule(lr){2-4} \cmidrule(lr){5-7} \cmidrule(lr){8-10}
\textbf{Based on $m_1$, Add New Agent} 
& $+m_2$ & $+m_3$ & $+m_4$ 
& $+m_2$ & $+m_3$ & $+m_4$ 
& $+m_2$ & $+m_3$ & $+m_4$ \\
\midrule
No Fusion & 74.8 & 74.8 & 74.8 & 60.6 & 60.6 & 60.6 & / & / & / \\
Late Fusion & 77.5 & 83.3 & 83.4 & 59.9 & 68.5 & 68.5 &  / & /& / \\
\midrule
%F-Cooper \cite{b32} & 0.778 & 0.742 & 0.761 & 0.628 & 0.517 & 0.494 & 30.70 & 39.59 & 49.22\\
%DiscoNet \cite{b33} & 0.798 & 0.833 & 0.830 & 0.653 & 0.682 & 0.695 &  30.80 & 39.67 & 49.29\\
%AttFusion \cite{b26} & 0.796 & 0.821 & 0.813 & 0.635 & 0.685 & 0.659 & 30.78 & 39.60 & 49.22\\
%V2X-ViT \cite{b34} & 0.822 & 0.888 & 0.882 & 0.655 & 0.765 & 0.753 &  36.17 & 44.99 & 54.62\\
CoBEVT~\cite{b35} & 82.2 & 88.5 & 88.5 & 67.1 & 74.2 & 74.2 & 33.24 & 42.05 & 51.68\\
%HM-ViT \cite{b9} & 0.813 & 0.871 & 0.876 & 0.646 & 0.743 & 0.755 & 47.71 & 65.08 & 83.34\\
HEAL (reported by~\cite{b12}) & 82.6 & 89.2 & 89.4 & 72.6 & 81.2 & 81.3 & 15.01 & 1.08 & 1.91   \\
HEAL (trained with ours configurations)   & 87.2 & 91.3 & 90.9& 78.3& 84.8& 84.5& 15.01 & 1.08 & 1.91\\
\textbf{\textsc{Faster-HEAL} (ours)} &\textbf{89.3} & \textbf{93.0} &\textbf{92.7} & \textbf{79.7} & \textbf{86.3} & \textbf{85.7} & \textbf{ 0.113}  & \textbf{0.113}& \textbf{0.113} \\
Improvement w.r. to HEAL & \textbf{$\uparrow 2.1$} & \textbf{$\uparrow 1.7$ } & \textbf{ $\uparrow 1.8$ } & \textbf{$\uparrow 1.4$ } & \textbf{$\uparrow 1.5$ } & \textbf{$\uparrow 1.2$ } & \textbf{($\downarrow 99.2\%$)} & \textbf{($\downarrow 89.5\%$)}& \textbf{($\downarrow 94.1\%$)} \\
\bottomrule
\end{tabular}}
\end{table*}

%\begin{table*}[t]
%\centering
%\caption{Training computation cost comparison on OPV2V-H using $m_1$ as the collaborative base with three additional heterogeneous agents. \textsc{Faster-HEAL} achieves higher FLOPs utilization, improved training throughput, and lower peak memory usage compared to HEAL.}
%\label{tab:fast}

%\resizebox{\textwidth}{!}{
%\begin{tabular}{lccccccccc}
%\toprule
%\textbf{Metric} & \multicolumn{3}{c}{\textbf{Train Throughput (\#/sec.)} $\uparrow$} & \multicolumn{3}{c}{\textbf{FLOPs (T\#/sec.)} $\uparrow$} & %\multicolumn{3}{c}{\textbf{Peak Memory (GB)} $\downarrow$} \\
%\cmidrule(lr){2-4} \cmidrule(lr){5-7} %\cmidrule(lr){8-10}
%\textbf{Based on $m_1$, Add New Agent} 
%& $+m_2$ & $+m_3$ & $+m_4$ 
%& $+m_2$ & $+m_3$ & $+m_4$ 
%& $+m_2$ & $+m_3$ & $+m_4$ \\
%\midrule
%HEAL \cite{b12} & 3.86 & 16.81 & 5.45 & 3.45  & 7.51 & 2.98  & 3.12 & 1.68 & 2.17 \\
%\textbf{\textsc{Faster-HEAL} (ours)} & \textbf{4.63} &\textbf{20.10} & \textbf{5.71}  & \textbf{4.31} & \textbf{8.99 } & \textbf{3.33} &\textbf{1.48 }& \textbf{1.40} & \textbf{1.42} \\
%Improvement w.r. to HEAL & \textbf{$\uparrow1.20\times$} & \textbf{$\uparrow1.19\times$ } & \textbf{ $\uparrow1.05\times$ } & \textbf{$\uparrow1.25\times$ } & \textbf{$\uparrow1.19\times$ } & \textbf{$\uparrow1.13\times$ } & \textbf{($\downarrow32.4\%$)} & \textbf{($\downarrow16.7\%$)}& \textbf{($\downarrow34.0\%$)} \\
%\bottomrule
%\end{tabular}}
%\end{table*}
\begin{table}[t]
\centering
\caption{Training computation cost comparison on OPV2V-H using $m_1$ as the collaborative base, summed for new heterogeneous agents ($m_2 , m_3 , m_4$). TT = training throughput (samples/sec.), TF = TFLOPs  (TFLOPs/sec.), and PM = peak memory usage.} 
%\textsc{Faster-HEAL} achieves higher FLOPs utilization, improved training throughput, and lower peak memory usage compared to HEAL.}
\label{tab:fast}
\resizebox{0.85\columnwidth}{!}{
\begin{tabular}{lccc}
\toprule
\textbf{Metric} & TT $\uparrow$ & TF $\uparrow$ & PM (GB) $\downarrow$ \\
%\cmidrule(lr){2-4} \cmidrule(lr){5-7} \cmidrule(lr){8-10}
%\textbf{Based on $m_1$, Add New Agent} 
%& $+m_2$ & $+m_3$ & $+m_4$ 
%& $+m_2$ & $+m_3$ & $+m_4$ 
%& $+m_2$ & $+m_3$ & $+m_4$ \\
\midrule
HEAL~\cite{b12} &26.12 & 13.94&7.0 \\
\textbf{\textsc{Faster-HEAL} (ours)} &  \textbf{30.44}&\textbf{16.63} &\textbf{4.3} \\
Improvement & \textbf{$\uparrow1.16\times$ } & \textbf{$\uparrow1.19\times$ } & \textbf{($\downarrow38.6\%$)} \\
\bottomrule
\end{tabular}}
\end{table}
%We further evaluate the robustness of \textsc{Faster-HEAL} under localization noise, which is a common challenge in collaborative perception due to imperfect pose estimation and GPS drift. To simulate pose errors, we inject zero-mean Gaussian noise $N(0,\sigma^2)$ into both the position and yaw angle of each agent. As shown in Figure~ \ref{fig:error}, \textsc{Faster-HEAL} consistently outperforms SOTA frameworks across different noise levels, exhibiting a slower performance degradation and demonstrating improved robustness to localization errors.

%\begin{figure}
   % \centering
   % \includegraphics[width=1\linewidth]{ap50_ap70_comparison.pdf}
 %   \caption{Effect of localization error on detection performance. Zero-mean Gaussian noise $N(0,\sigma^2)$ is applied to agent position and yaw angle. \textsc{Faster-HEAL} exhibits more robust performance than SOTA frameworks across increasing noise levels.}

  %  \label{fig:error}
%\end{figure}
\subsection{Ablation Study}
%We analyze the contribution of each trainable component in stage two of \textsc{Faster-HEAL}. For fairness, we fix the decomposition rank to $R=8$ across all cases and compare freezing versus fine-tuning the encoder, BEV backbone, and visual prompts. As shown in Table~\ref{tab:ablation-study}, fine-tuning only the visual prompts provides the largest performance gain, most effectively addressing the domain gap while requiring just $340\text{k}$ parameters. In contrast, retraining the encoder or BEV backbone alongside prompt tuning yields no additional benefits, but instead causes overfitting and slower convergence. These results demonstrate that lightweight prompt fine-tuning offers the best trade-off between accuracy and efficiency, validating our design choice.
We analyze the contribution of each trainable component in stage~2 of \textsc{Faster-HEAL}. We fix the decomposition rank at $R=8$ and compare freezing versus fine-tuning the encoder, BEV backbone, and \textsc{LIFT}. As shown in Table~\ref{tab:ablation-study}, fine-tuning only the \textsc{LIFT} yields the largest performance gain, most effectively addressing the domain gap, and uses only $340\text{k}$ parameters. In contrast, retraining the encoder or BEV backbone alongside \textsc{LIFT} brings no extra benefits and instead causes overfitting and slower convergence. This confirms that lightweight prompt fine-tuning offers the best accuracy–efficiency trade-off and validates our design choice.
\begin{table}[t]
\centering
\caption{Ablation study on stage~2 training components for new agent adaptation ($m_2$, $m_3$, $m_4$), with $m_1$ as the collaborative base. We compare freezing or tuning the encoder, BEV backbone (BEV), and LIFT. The aligner is trained in all cases.}
%$\#$TP = number of trainable parameters, TT = training throughput, TF = TFLOPs, and PM = peak memory usage. Results show that visual prompt fine-tuning yields the best performance–efficiency trade-off. The results are reported for $m_1 + m_2 + m_3 + m_4$.}
\label{tab:ablation-study}
\resizebox{0.9\columnwidth}{!}{
%\begin{tabular}{cccccccccc}
%\toprule
%\textbf{Model} & \textbf{Encoder} & \textbf{BEV Backbone} & \textbf{Visual Prompt} & \textbf{AP50 $\uparrow$} & \textbf{AP70 $\uparrow$} & {\textbf{\#TP (M)} $\downarrow$}  & {\textbf{TT (\#/sec.)} $\uparrow$} & {\textbf{TF (\#/sec.)} $\uparrow$} & {\textbf{PM (GB)} $\downarrow$}  \\
%\midrule
%HEAL & \checkmark & \checkmark & \xmark & 0.909 & 0.845 & 18 &  26.15 & 13.95 & 6.97 \\
%Case 1 & \checkmark & \checkmark & \checkmark& 0.896 & 0.831 & 18 & 24.00 & 13.28 & 7.45 \\
%Case 2 & \checkmark & \faLock & \checkmark & 0.922 & 0.853 & 17.2 & 27.45 & 14.76 & 7.44 \\
%Case 3 & \faLock & \checkmark & \checkmark & 0.907 & 0.836 & 1.14 & 28.74 & 16.32 & 4.76\\
%\textbf{Faseter-HEAL(ours)} & \faLock & \faLock & \checkmark & \textbf{0.927} & \textbf{0.857} & \textbf{0.34} & \textbf{30.43} & \textbf{16.65} &\textbf{4.30} \\
%\bottomrule
%\end{tabular}
%\end{table*}
\begin{tabular}{cccc}
\toprule
\textbf{Training Components} &  \textbf{AP@0.5(\%) $\uparrow$} & \textbf{AP@0.7(\%) $\uparrow$} & {\textbf{\#TP (M)} $\downarrow$} \\
\midrule
Encoder + BEV (HEAL) & 90.9 & 84.5 & 18\\
Encoder + BEV + LIFT & 89.6 & 83.1 & 18 \\
Encoder + LIFT & 92.2 & 85.3 & 17.2\\
BEV + LIFT& 90.7 & 83.6 & 1.14\\
\textbf{LIFT (ours)} &  \textbf{92.7} & \textbf{85.7} & \textbf{0.34} \\
\bottomrule
\end{tabular}}
\end{table}
%We further evaluate the effect of the decomposition rank $R$ in CP decomposition on model performance. As shown in Equation~\ref{equ:cp}, each element of the prompt $P$ is reconstructed by summing over $R$. While the number of trainable parameters varies with $R$ (between 3\% and 9\%), the effect of this change on training speed is negligible. To ensure fairness, we keep all configurations fixed and vary only $R$. As shown in Table~\ref{tab:ablation-study1}, $R=8$ achieves the best trade-off between performance and efficiency: $R=4$ leads to underfitting, while $R=16$ and $R=32$ result in overfitting and slower convergence.

We further evaluate the effect of the PARAFAC decomposition rank $R$ on model performance. As in Eq.~\ref{equ:cp}, each prompt element $P$ is reconstructed by summing over $R$. Hence, the number of trainable parameters varies with $R$ (varies 3\%-9\%), while its impact on training speed is negligible. For fairness, we fix all other configurations and vary only $R$. Table~\ref{tab:ablation-study1} shows that $R=8$ provides the best performance–efficiency trade-off, where $R=4$ underfits, and $R=16$ and $R=32$ lead to overfitting and slower convergence.
\begin{table}[t]
\centering
\caption{Effect of decomposition rank $R$ in PARAFAC decomposition on new-agent adaptation. We report detection accuracy (AP@0.5, AP@0.7) and the number of trainable parameters (\#TP). Results show that $R=8$ provides the best trade-off The results are reported for $m_1 + m_2 + m_3 + m_4$.}
%, with $R=4$ underfitting and $R=16$ or $R=32$ leading to slower convergence. The results are reported for $m_1 + m_2 + m_3 + m_4$.}
\label{tab:ablation-study1}
\resizebox{0.80\columnwidth}{!}{
\begin{tabular}{cccc}
\toprule
\textbf{R} & \textbf{AP@0.5(\%) $\uparrow$} & \textbf{AP@0.7(\%) $\uparrow$} & {\textbf{\#TP (M)} $\downarrow$} \\
\midrule
4 & 91.6 & 84.0 & \textbf{0.335} \\
\textbf{8} & \textbf{92.7} & \textbf{85.7} & 0.340 \\
16 & 89.7 & 82.1 & 0.352 \\
32 & 91.4& 84.0& 0.373 \\
64 &  90.2 &83.0 & 0.416 \\
\bottomrule
\end{tabular}}
\end{table}
\section{Conclusion}
%In this paper, we introduced \textbf{Faster-HEAL}, a novel framework for open heterogeneous collaborative perception. By fine-tuning lightweight visual prompts \textbf{LIFT} in a unified feature space, our method achieves state-of-the-art detection performance while requiring only a fraction of the trainable parameters compared to prior approaches, thereby preserving computational efficiency. Moreover, it safeguards the privacy of participating agents since training does not require access to the internal architecture or parameters of new agent models, relying solely on their shared intermediate features. Extensive experiments on the OPV2V-H dataset demonstrate the effectiveness of \textsc{Faster-HEAL} in mitigating the feature domain gap caused by heterogeneous sensors and models, enabling fast adaptation without retraining entire perception stacks. These results represent an important step toward practical, scalable, and privacy-preserving collaborative perception for real-world autonomous driving systems.
In this paper, we introduced \textbf{Faster-HEAL}, a novel framework for open heterogeneous collaborative perception that fine-tunes lightweight visual prompts (\textbf{LIFT}) in a unified feature space. Our method achieves state-of-the-art detection performance with only a fraction of the trainable parameters of prior methods, preserving computational efficiency and privacy, since training relies solely on shared intermediate features and does not require access to internal architectures or model parameters of the new agent. Experiments on OPV2V-H demonstrate that \textsc{Faster-HEAL} mitigates the feature domain gap caused by heterogeneous sensors/models effectively, enabling fast adaptation without retraining entire perception stacks and representing a practical, scalable privacy-preserving CP framework in real-world autonomous driving. Extending the evaluation to additional CP datasets/baselines, and more diverse sensor/model/noise configurations, is left for the future.

\noindent
\textbf{Acknowledgment} This work was supported by a Department of Transportation/Federal Railroad Administration grant.
%\section*{Acknowledgment}
%The preferred spelling of the word ``acknowledgment'' in America is without an ``e'' after the ``g''. Avoid the stilted expression ``one of us (R. B.  G.) thanks $\ldots$''. Instead, try ``R. B. G. thanks$\ldots$''. Put sponsor acknowledgments in the unnumbered footnote on the first page.

%\section*{References}

%Please number citations consecutively within brackets \cite{b1}. The sentence punctuation follows the bracket \cite{b2}. Refer simply to the reference  number, as in \cite{b3}---do not use ``Ref. \cite{b3}'' or ``reference \cite{b3}'' except at the beginning of a sentence: ``Reference \cite{b3} was the first $\ldots$'' Number footnotes separately in superscripts. Place the actual footnote at the bottom of the column in which it was cited. Do not put footnotes in the abstract or reference list. Use letters for table footnotes. Unless there are six authors or more give all authors' names; do not use ``et al.''. Papers that have not been published, even if they have been submitted for publication, should be cited as ``unpublished'' \cite{b4}. Papers that have been accepted for publication should be cited as ``in press'' \cite{b5}. Capitalize only the first word in a paper title, except for proper nouns and element symbols. For papers published in translation journals, please give the English citation first, followed by the original foreign-language citation \cite{b6}.

\end{document}